%% file: main.tex
\documentclass[a4paper]{article}
\usepackage{graphicx}
\usepackage{twocolceurws}
\usepackage{rotating}
\usepackage{multirow}
\usepackage{subcaption}
\usepackage{url}

\title{More is not Always Better: The Negative Impact of A-box Materialization on RDF2vec Knowledge Graph Embeddings}

\author{Andreea Iana \and Heiko Paulheim}

\institution{Data and Web Science Group\\University of Mannheim, Germany\\
\texttt{aiana@mail.uni-mannheim.de, heiko@informatik.uni-mannheim.de}}

\begin{document}
\maketitle

\begin{abstract}
RDF2vec is an embedding technique for representing knowledge graph entities in a continuous vector space.
In this paper, we investigate the effect of materializing implicit A-box axioms induced by subproperties, as well as symmetric and transitive properties.
While it might be a reasonable assumption that such a materialization before computing embeddings might lead to better embeddings, we conduct a set of experiments on DBpedia which demonstrate that the materialization actually has a negative effect on the performance of RDF2vec.
In our analysis, we argue that despite the huge body of work devoted on completing missing information in knowledge graphs, such missing implicit information is actually a \emph{signal}, not a \emph{defect}, and we show examples illustrating that assumption.
\end{abstract}

\section{Introduction}

RDFvec \cite{ristoski2016rdf2vec} was originally conceived for exploiting knowledge graphs in data mining. Since most popular data mining tools require a feature vector representation of records, various techniques have been proposed for creating vector space representations from subgraphs, including adding datatype properties as features or creating binary features for types \cite{ristoski2014comparison}. Given the increasing popularity of the word2vec family of word embedding techniques \cite{mikolov2013efficient}, which learns feature vectors for words based on the context in which they appear, this approach has been proposed to be transferred to graphs as well. Since word2vec operates on (word) sequences, several approaches have been proposed which first turn a graph into sequences by performing random walks, before applying the idea of word2vec to those sequences. Such approaches include node2vec \cite{grover2016node2vec}, DeepWalk \cite{perozzi2014deepwalk}, 
%Wembedder \cite{nielsen2017wembedder}, 
and the aforementioned RDF2vec.

There is a plethora of work addressing the completion of knowledge graphs~\cite{paulheim2017knowledge}, i.e., the addition of missing knowledge. Since some knowledge graphs come with expressive schemas \cite{heist2020kgoverview} or exploit upper ontologies \cite{paulheim2015serving}, one such approach is the exploitation of explicit ontological knowledge. For example, if a property $p$ is known to be symmetric, a reverse edge $p(y,x)$ can be added to the knowledge graph for each edge $p(x,y)$ found. 

A straightforward assumption is that completing missing knowledge in a knowledge graph before computing node representations will lead to \emph{better} results. However, in this paper, we show that the opposite actually holds: completing the knowledge graph before computing an RDF2vec embedding actually leads to \emph{worse} results in downstream tasks.

%The rest of this paper is structured as follows. In section~\ref{sec:related}, we present some related works. Section~\ref{sec:experiments} describes our experimental setup and provides empirical evidence for the above observation. We discuss a number of causes and implications of those findings, also beyond RDF2vec, in section~\ref{sec:discussion}, and conclude with a summary and outlook on future work.

\section{Related Work}
\label{sec:related}
The base algorithm of RDF2vec uses random walks on the knowledge graph to produce sequences of nodes and edges. Those sequences are then fed into a word2vec embedding learner, i.e., using either the CBOW or the Skip-Gram method.

Since its original publication in 2016, several improvements for RDF2vec have been proposed. The main family of approaches for improving RDF2vec is to use alternatives for completely random walks to generate sequences. \cite{cochez2017biased} explores 12 variants of \emph{biased walks}, i.e., random walks which follow non-uniform probability distributions when choosing an edge to follow in a walk. Heuristics explored include, e.g., preferring successors with a high or low PageRank, preferring frequent or infrequent edges, etc. %The results, however, are inconclusive, and no globally best strategies could be identified by the authors.

In \cite{saeed2019extracting}, the authors explore the automatic identification of a relevant subset of edge types for a given class of entities. They show that restricting the graph for a class of entities at hand (e.g., movies) can outperform the results of pure RDF2vec.

While those works exploit merely knowledge graph internal signals (e.g., by computing PageRank over the graph), other works include external signals as well. For example, \cite{ahmad} shows that exploiting an external measure for the importance of an edge can lead to improved results over other biasing strategies. The authors utilize page transition probabilities obtained from server log files in Wikipedia to compute a probability distribution for creating the random walks. 

%Apart from those approaches for improving the qualitative performance of RDF2vec, approaches for making RDF2vec faster and easier to use have also been proposed. Those include a fast Python reimplementation with additional variants for walk generation\footnote{\url{https://github.com/IBCNServices/pyRDF2Vec}} and a REST service for retrieving pre-computed RDF2vec embeddings for different knowledge graphs \cite{portisch2020kgvec2go}.

A work that explores a similar direction to the one proposed in this paper is presented in \cite{mai2018support}. The authors analyze the information content of statements in a knowledge graph by computing how easily a statement can be predicted from the other statements in the knowledge graph. They show that translational embeddings can benefit from being tuned towards focusing on statements with a high information content.

\section{Experiments}
\label{sec:experiments}
To evaluate the effect of knowledge graph materialization on the quality of RDF2vec embeddings, we repeat the experiments on entity classification and regression, entity relatedness and similarity and document similarity introduced in \cite{ristoski2019rdf2vec}, and compare the results on the materialized and unmaterialized graphs.\footnote{Please note that the results on the unmaterialized graphs differ from those reported in \cite{ristoski2019rdf2vec}, since we use a more recent version of DBpedia in our experiments.}

\subsection{Experiment Setup}
For our experiments, we use the 2016-10 dump of DBpedia, which was the latest official release during the time at which the experiments were conducted.
For creating RDF2Vec embeddings, we use KGvec2go \cite{portisch2020kgvec2go} for computing the random walks, and the fast Python reimplementation of the original RDF2Vec code\footnote{\url{https://github.com/IBCNServices/pyRDF2Vec}} for training the RDF2Vec models\footnote{\url{https://github.com/andreeaiana/rdf2vec-materialization}
}.

Since the original DBpedia ontology provides information about subproperties, but does not define any symmetric, transitive, and inverse properties, we first had to enrich the ontology with such axioms. 
%To that end, we pursue two different strategies.

\subsubsection{Enrichment using Wikidata}
The first strategy is utilizing \texttt{owl:equivalentProperty} links to Wikidata \cite{vrandevcic2014wikidata}. We mark a property $P$ in DBpedia as symmetric if its Wikidata equivalent has a symmetric constraint in Wikidata\footnote{\url{https://www.wikidata.org/wiki/Q21510862}}, and we mark it as transitive if its Wikidata equivalent is an instance of the Wikidata class \emph{transitive property}\footnote{\url{https://www.wikidata.org/wiki/Q18647515}}. For a pair of properties $P$ and $Q$ in DBpedia, we mark them as inverse if their respective equivalent properties in Wikidata are defined as inverse of one another\footnote{\url{https://www.wikidata.org/wiki/Property:P1696}}.

% , and we mark them as subproperties if their respective equivalent properties in Wikidata are defined as subproperties\footnote{\url{https://www.wikidata.org/wiki/Property:P1647}}.

\subsubsection{Enrichment using DL-Learner}
The second strategy is applying DL-Learner \cite{lehmann2009dl} to learn additional symmetry, transitivity, and inverse axioms for enriching the ontology. After inspecting the results of DL-Learner, and to avoid false T-box axioms, we used thresholds of 0.53 for symmetric properties, and 0.45 for transitive properties. Since the list of pairs of inverse properties generated by DL-Learner contained quite a few false positives (e.g., \texttt{dbo:isPartOf} being the inverse of \texttt{dbo:countySeat} as the highest scoring result), we manually filtered the top results and kept 14 T-box axioms which we rated as correct.

\subsubsection{Materializing the Enriched Graphs}
In both cases, we identify a number of inverse, transitive, and symmetric properties, as shown in Table~\ref{tab:datasets}. The symmetric properties identified by the two approaches highly overlap, while the inverse and transitive properties identified differ a lot.

With the enriched ontology, we infer additional A-box axioms on DBpedia. We use two settings, i.e., all subproperties plus (a) all inverse, transitive, and symmetric properties found using mappings to Wikidata, and (b) all plus all inverse, transitive, and symmetric properties found with DL-Learner.

The inferring of additional A-box axioms was done in iterations. In each iteration, additional A-box axioms were created for symmetric, transitive, inverse, and subproperties. Using this iterative approach, chains of properties could also be respected. For example, from the axioms
\begin{verbatim}
Cerebellar_tonsil isPartOfAnatomicalStructure 
   Cerebellum .
Cerebellum isPartOfAnatomicalStructure
   Hindbrain .
\end{verbatim}
and the two identified T-box axioms
\begin{verbatim}
isPartOf a owl:TransitiveProperty .
isPartOfAnatomicalStructure rdfs:subPropertyOf 
   isPartOf .
\end{verbatim}
the first iteration adds
\begin{verbatim}
    Cerebellar_tonsil isPartOf Cerebellum .
    Cerebellum isPartOf Hindbrain .
\end{verbatim}
whereas the second iteration adds
\begin{verbatim}
    Cerebellar_tonsil isPartOf Hindbrain .
\end{verbatim}
The materialization process is terminated once no further axioms are added. This happens after two iterations for the dataset enriched with Wikidata, and three iterations for the dataset enriched with DL-Learner. The size of the resulting datasets is shown in Table~\ref{tab:datasets}. 

%\begin{table*}[ht]
%    \caption{Materialization Rules}
%    \label{tab:materialization_rules}
%    \scriptsize
%    \centering
%    \begin{tabular}{l|l|l}
%         T-box axiom & Observed A-box axiom(s) & Added A-box axiom \\
%         \hline
%         \texttt{p rdfs:subPropertyOf q .} & \texttt{x p y .} & \texttt{x q y .} \\
%         \texttt{p owl:inversePropertyOf q .} & \texttt{x p y .} & \texttt{y q x .} \\
%         \texttt{p a owl:SymmetricProperty .} & \texttt{x p y .} & \texttt{y p x .} \\
%         \texttt{p a owl:TransitiveProperty .} & \texttt{x p y . y p z .} & \texttt{x p z .} \\
%    \end{tabular}
%\end{table*}

%Fig.~\ref{fig:top10properties} depicts the properties most frequently added to the A-box by the different enrichment processes. It can be observed that in both cases, the majority of axioms is added due to the transitivity of \texttt{dbo:isPartOf}.

\begin{table*}[ht]
    \caption{Enriched DBpedia Versions Used in the Experiments. The upper part of the table depicts the number of T-box axioms identified with the two enrichment approaches, the lower part depicts the number of A-box axioms creatd by materializing the A-box according to the additional T-box axioms.}
    \label{tab:datasets}
    \scriptsize
    \centering
    \begin{tabular}{l|r|r|r}
         &  Original & Enriched Wikidata & Enriched DL-Learner \\
        \hline
        T-box subproperties & 75 & 0 &  0 \\
        T-box inverse properties & 0 & 8 & 14 \\ 
        T-box transitive properties  & 0 & 7 & 6 \\
        T-box symmetric properties & 0 & 3 & 7 \\
         \hline
         A-box subproperties & -- & 122,491 & 129,490 \\
         A-box inverse properties & -- & 44,826 & 159,974 \\
         A-box transitive properties & -- & 334,406 & 415,881 \\
         A-box symmetric properties & -- & 4,115 & 35,885 \\
         \hline
         No. of added triples & -- & 505,838 & 741,230 \\
         No. of total triples & 50,000,412 & 50,506,250 & 50,741,642 \\
    \end{tabular}
\end{table*}

%\begin{figure}[h]
%    \centering
%    \begin{subfigure}[c]{0.45\textwidth}
%        \includegraphics[width=\textwidth]{figures/top10_wikidata_cropped.pdf}
%        \caption{Wikidata}
%    \end{subfigure}
%    ~
%    \begin{subfigure}[c]{0.45\textwidth}
%        \includegraphics[width=\textwidth]{figures/top10_dllearner_cropped.pdf}
%        \caption{DLLearner}
%    \end{subfigure}
%    \caption{10 properties with most axioms added to the A-box}
%    \label{fig:top10properties}
%\end{figure}

\subsection{Training RDF2vec Embeddings}
On all three graphs (\emph{Original}, \emph{Enriched Wikidata}, and \emph{Enriched DL-Learner}), experiments were conducted in the same fashion as in \cite{ristoski2019rdf2vec}. The RDF2vec approach extracts sequences of nodes and properties by performing random walks from each node. Following~\cite{ristoski2016rdf2vec}, we started 500 random graph walks of depth 4 and 8 from each node.

The resulting sequences are then used as input to word2vec. Here, two variants exist, i.e., CBOW and Skip-Gram (SG), where SG consistently yielded better results in~\cite{ristoski2016rdf2vec}, so we used the SG to compute embeddings vectors with a dimensionality of 200 and 500. Following~\cite{ristoski2016rdf2vec}, the parameters chosen for word2vec were window size = 5, no. of iterations = 10, and negative sampling with no. of samples = 25. The code and data used for the experiments are available online.\footnote{\url{https://github.com/andreeaiana/rdf2vec-materialization}
}

\subsubsection{Experiments Conducted on the Enriched Graphs}
This results in 12 different embeddings to be compared against each other. For evaluation, we use the evaluation framework provided in \cite{pellegrino2019configurable}. The tasks to evaluate were
\begin{enumerate}
    \item Regression: five regression datasets where an external variable not contained in DBpedia is to be predicted for a set of entities (cities, universities, companies, movies, and albums);
    \item Classification: five classification datasets derived from the aforementioned regression dataset by discretizing the target variable;
    \item Entity relatedness and entity similarity, based on the KORE50 dataset; and % \cite{hoffart2012kore}
    \item Document similarity, based on the LP50 dataset,
    %\cite{lee2005empirical}, 
    where the similarity of two documents is computed from the pairwise similarities of entities identified in the texts.
\end{enumerate}

The experimental protocol in the framework used for evaluation is defined as follows \cite{pellegrino2019configurable}:

For regression and classification, three (linear regression, k-NN, M5 rules) resp. four (Naive Bayes, C4.5 decision tree, k-NN, Support Vector Machine) are used and evaluated using 10-fold cross validation. k-NN is used with k=3; for SVM, the parameter C is varied between $10^{-3}, 10^{-2}, 0.1, 1, 10, 10^2, 10^3$, and the best value is chosen. All other algorithms are run in their respective standard configurations.\footnote{\url {https://github.com/mariaangelapellegrino/Evaluation-Framework/blob/master/doc/Classification.md}}$^,$\footnote{\url{https://github.com/mariaangelapellegrino/Evaluation-Framework/blob/master/doc/Regression.md}}

For entity relatedness and similarity, the task is to rank a list of entities w.r.t. a main entity. Here, the entities are ranked by cosine similarity between the main entity's and the candidate entities' RDF2vec vectors.\footnote{\url{https://github.com/mariaangelapellegrino/Evaluation-Framework/blob/master/doc/EntityRelatedness.md}}

For the document similarity task, the similarity of two documents $d_1$ and $d_2$ is computed by comparing all entities mentioned in $d_1$ to all entities mentioned in $d_2$ using the metric above. For each entity in each document, the maximum similarity to an entity in the other document is considered, and the similarity of $d_1$ and $d_2$ is computed as the average of those maxima.\footnote{\url{https://github.com/mariaangelapellegrino/Evaluation-Framework/blob/master/doc/DocumentSimilarity.md}}

\subsection{Results on Different Tasks}
The first step of experiments are regression and classification, with the results depicted in Tables~\ref{tab:regression} and~\ref{tab:resultsClassification}. For the regression task, we can observe that the best result for each combination of a task and RDF2vec configuration (depth of walks, and dimensionality) is achieved on the unmaterialized graph in 15 out of 20 cases, with linear regression or KNN delivering the best results. If we consider all combinations of a task, an embedding, and a learner, the unmaterialized graph yields better results in 39 out of 60 cases.

The observations for classification are similar. For 19 out of 20 combinations of a task and an RDF2vec configuration, the best results are obtained on the original, unmaterialized graphs, most often with an SVM. If we consider all combinations of a task, an embedding, and a learner, the unmaterialized graph yields better results in 60 out of 80 cases.

Moreover, if we look at \emph{how much} the results degrade for the materialized graphs, we can observe that the variation is much stronger for the longer walks of depth 8 than the shorter walks of depth 4.
\input{table_classification_regression}

\input{table_similarity}

The observations on the other tasks are similar. For entity similarity, we see that better results are achieved on the unmaterialized graphs in 16 out of 20 cases, and in all of the four overall considerations. As far as entity relatedness is concerned, the results on the unmaterialized graphs are better in 13 out of 20 cases, as well as in all four overall considerations. It is noteworthy that only in three out of ten cases -- enriching the IT companies test set with DL-Learner and Wikidata, and enriching the Hollywood celebrities test set with Wikidata -- the degree of the entities at hand changes. This hints at the effects (both positive and negative) being mainly caused by information being added to the entities connected to the entities at hand (e.g., the company producing a video game), which is ultimately reflected in the walks.

\input{table_document}
Finally, for document similarity, we see a different picture. Here, the results on the unmaterialized graphs are always outperformed by those obtained on the materialized graphs, regardless of whether the embeddings were computed on the shorter or longer walks. The exact reason for this observation is not known. One observation, however, is that the entities in the LP50 dataset have by far the largest average degree (2,088, as opposed to only 18 and 19 for the MetacriticMovies and MetacriticAlbums dataset, respectively). Due to the already pretty large degree, it is less likely that the materialization skews the distributions in the random walks too much, and, instead, actually adds meaningful information. Another possible reason is that the entities in LP50 are very diverse (as opposed to a uniform set of cities, movies, or albums), and that in such a diverse dataset, the effect of materialization is different, as it tends to add heterogeneous rather than homogeneous information to the walks.

\subsection{A Closer Look at the Generated Walks}

\begin{figure*}[h!]
    \centering
    \begin{subfigure}[c]{0.31\textwidth}
        \includegraphics[width=\textwidth]{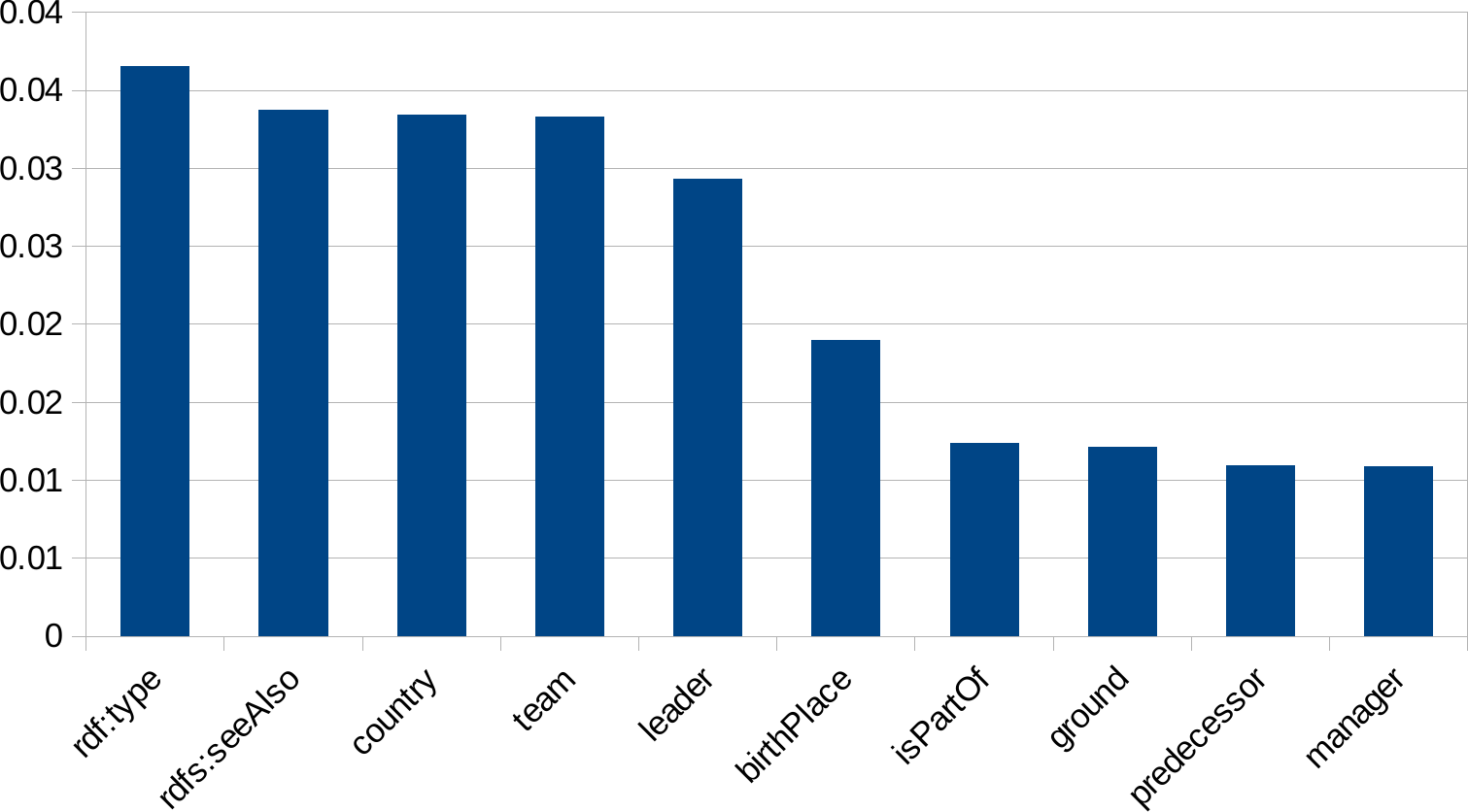}
        \caption{depth=4, original}
    \end{subfigure}
    ~
    \begin{subfigure}[c]{0.31\textwidth}
        \includegraphics[width=\textwidth]{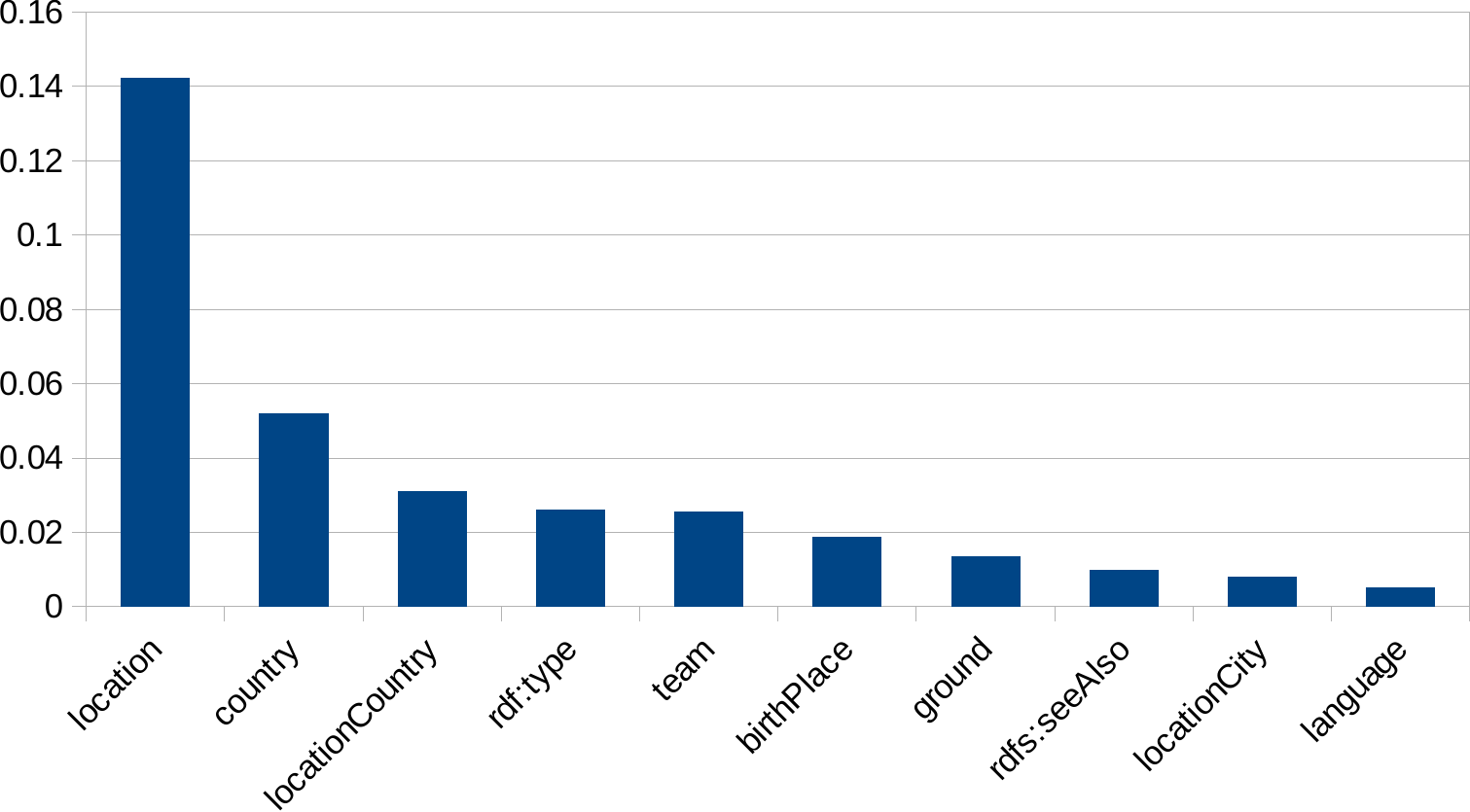}
        \caption{depth=4, Wikidata}
    \end{subfigure}
    ~
    \begin{subfigure}[c]{0.31\textwidth}
        \includegraphics[width=\textwidth]{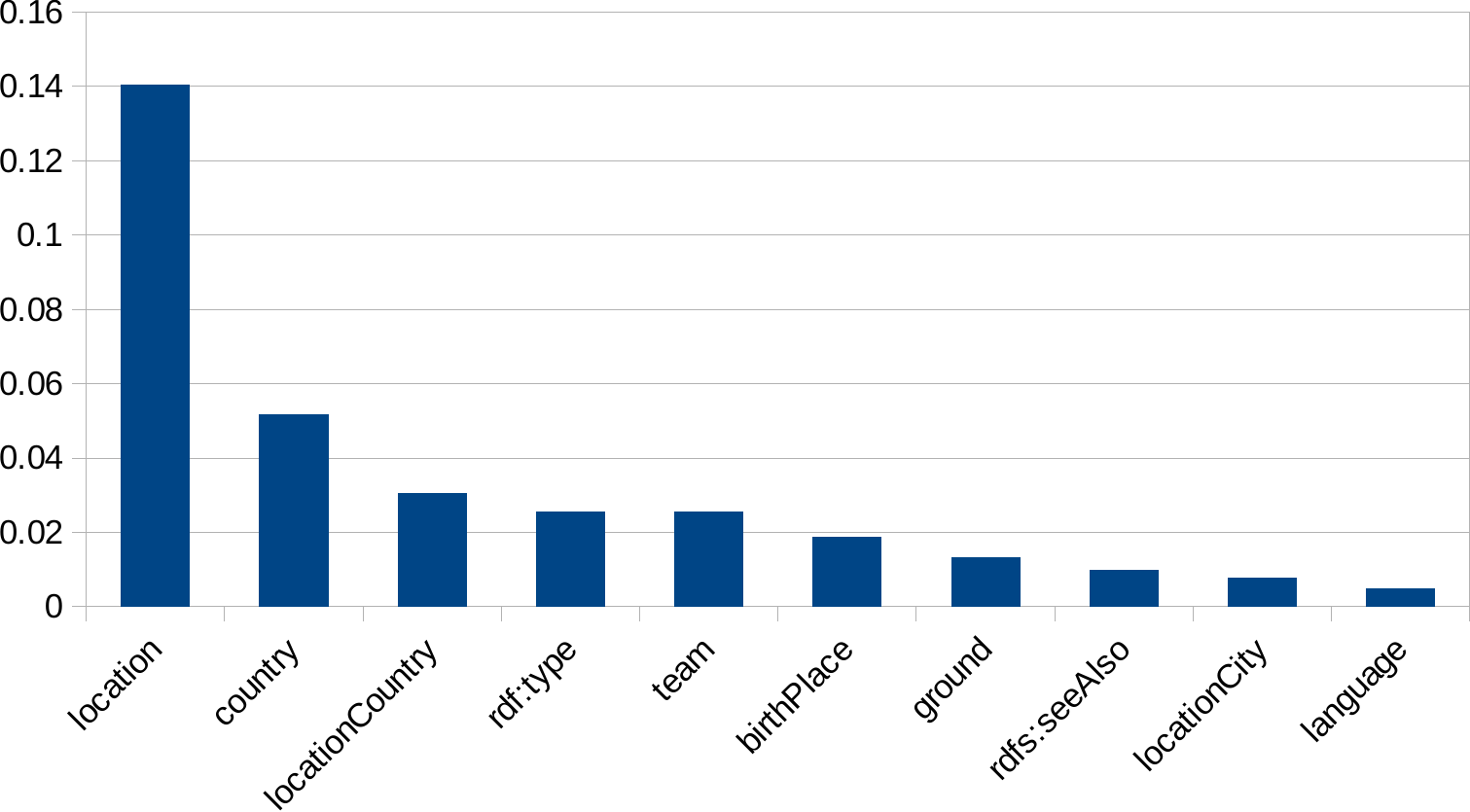}
        \caption{depth=4, DL-Learner}
    \end{subfigure}
    \\\vspace{0.25cm}
    \begin{subfigure}[c]{0.31\textwidth}
        \includegraphics[width=\textwidth]{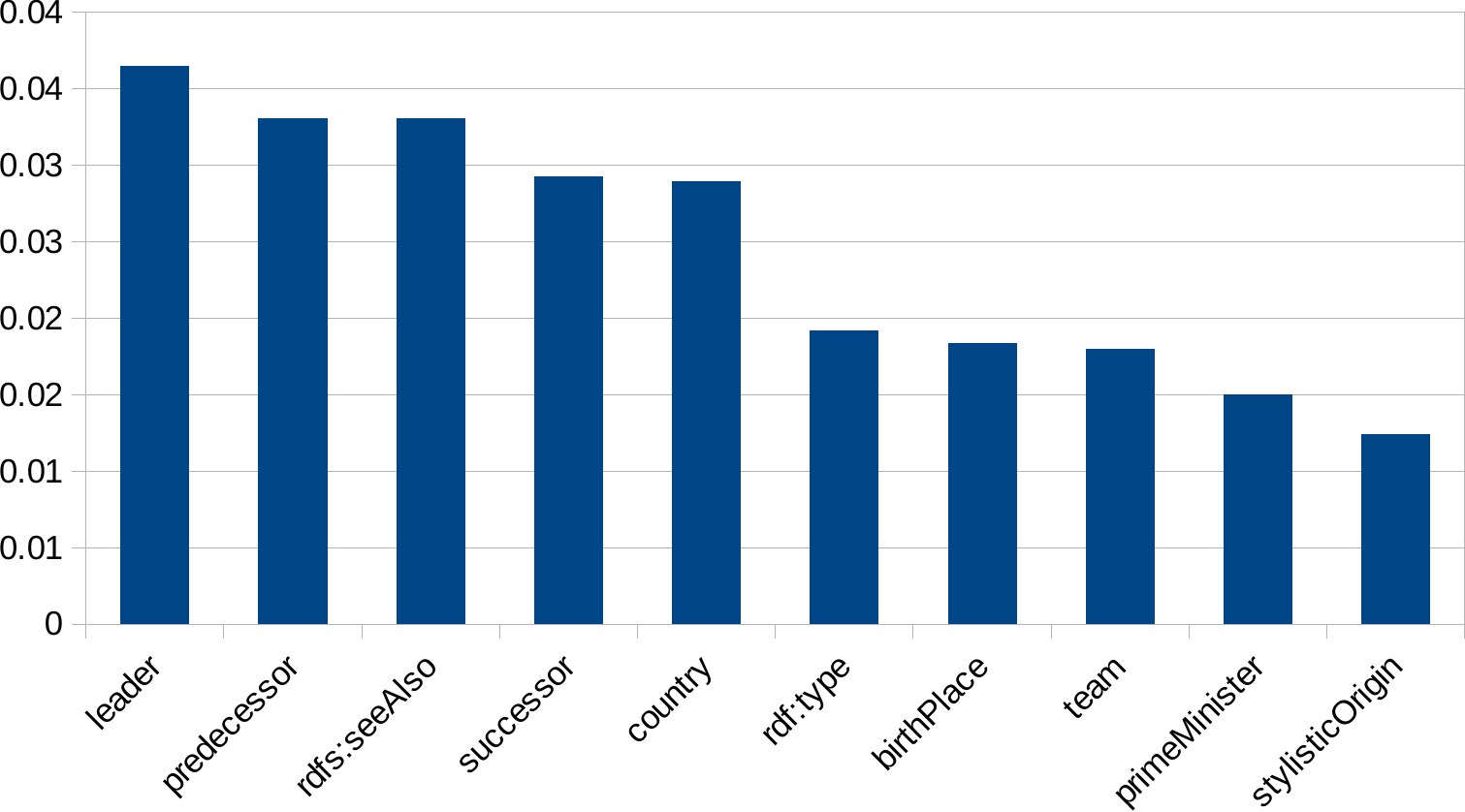}
        \caption{depth=8, original}
    \end{subfigure}
    ~
    \begin{subfigure}[c]{0.31\textwidth}
        \includegraphics[width=\textwidth]{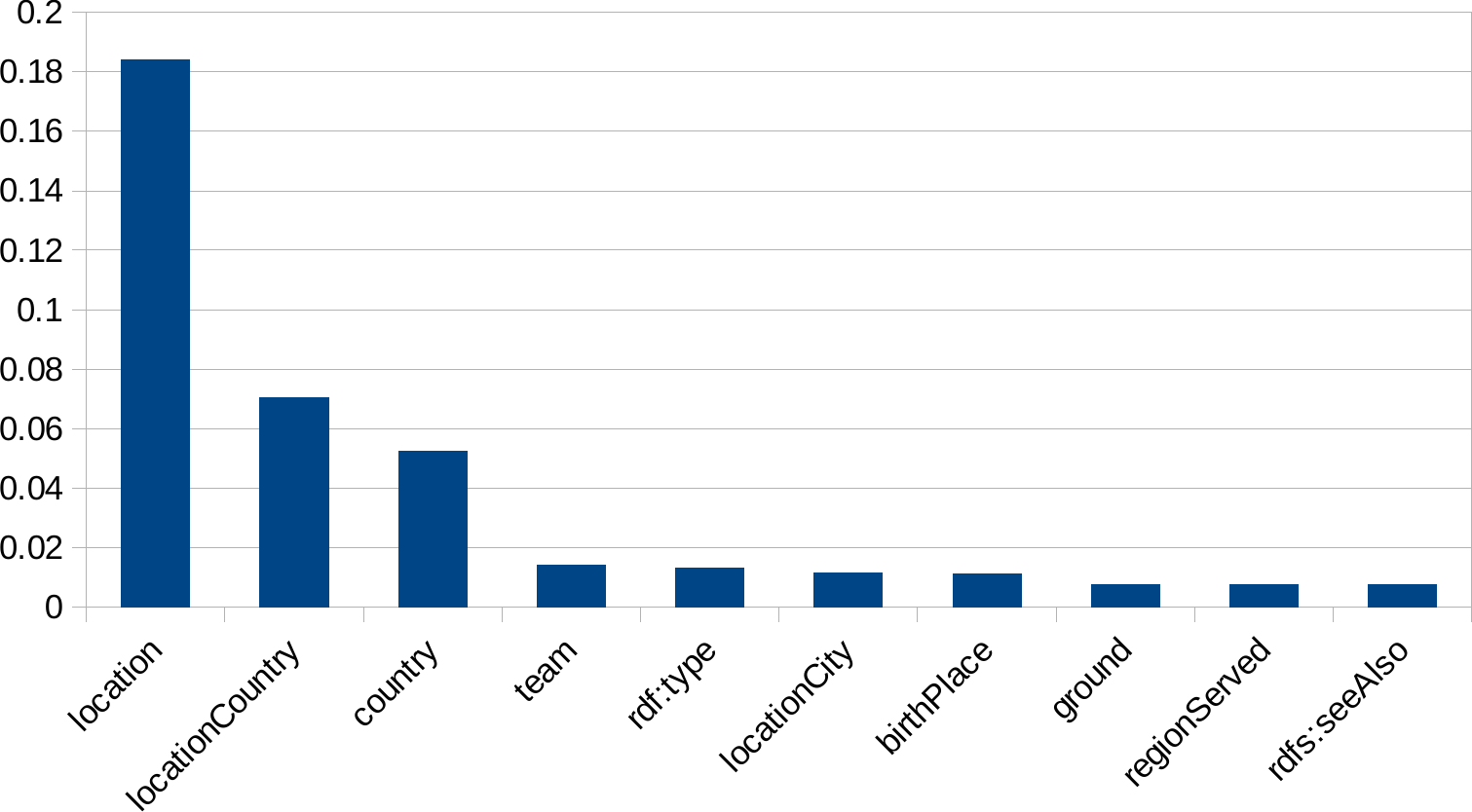}
        \caption{depth=8, Wikidata}
    \end{subfigure}
    ~
    \begin{subfigure}[c]{0.31\textwidth}
        \includegraphics[width=\textwidth]{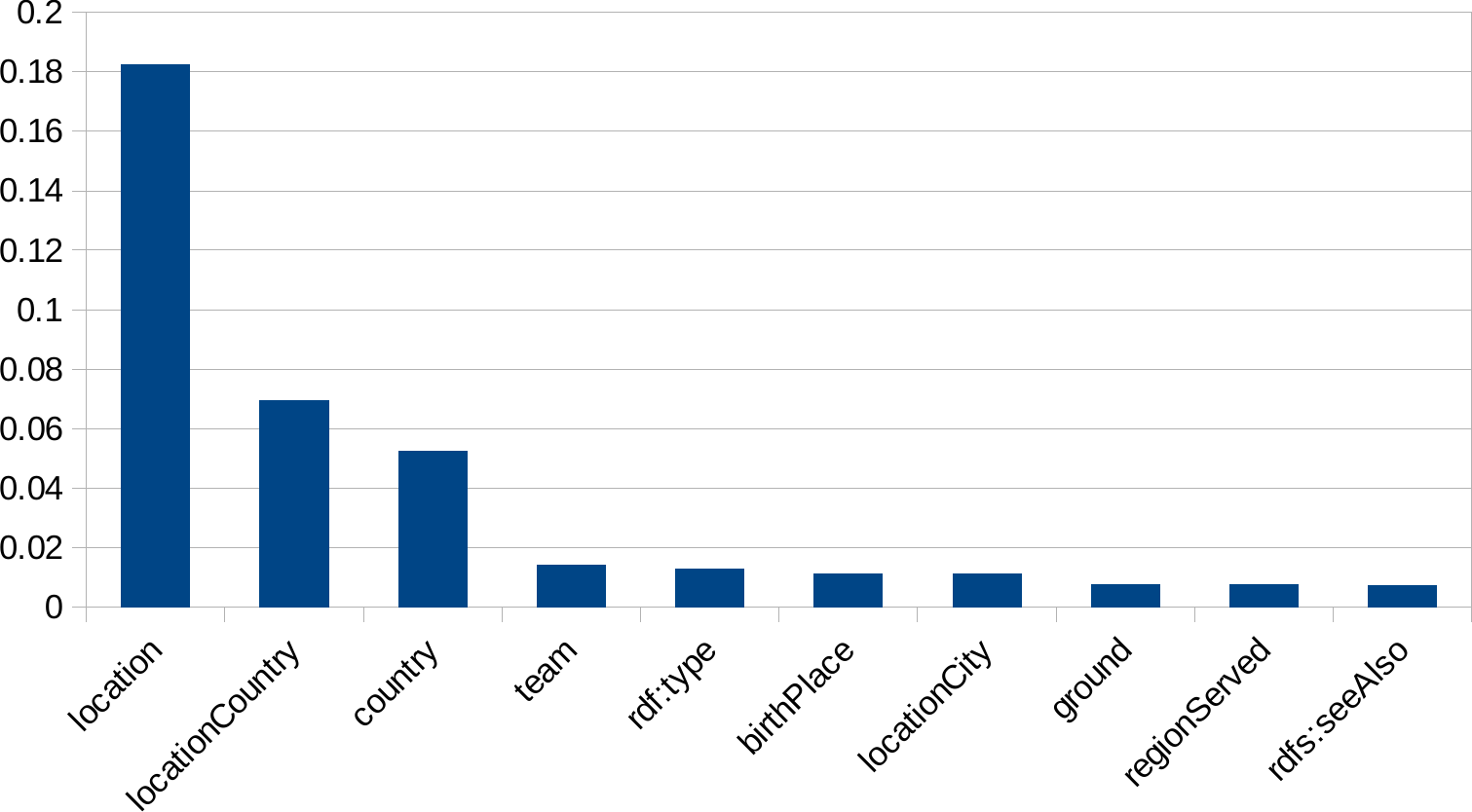}
        \caption{depth=8, DL-Learner}
    \end{subfigure}
    \caption{Distribution of top 10 properties in the generated walks}
    \label{fig:distribution_walks}
\end{figure*}

In order to analyze the findings above, we first tried to correlate the findings with the actual change on the entities in the respective test sets. However, there is no clear trend which can be identified. For example, in the classification and regression cases, the dataset which is most negatively impacted by materialization, i.e., the Metacritic Albums dataset, has the lowest change in its instances' degree (the avg. degree of the instances changes by 0.003\% and 0.007\% with the Wikidata and the DL-Learner enrichment, respectively). On the other hand, the increase in the degree of the instances on the cities dataset is much stronger (1.03\% and 1.04\%), while the decrease of the predictive models on that dataset is comparatively low.

We also took a closer look at the generated random walks on the different graphs. To that end, we computed distributions of all properties occurring in the random graph walks, for both strategies and for both depths of 4 and 8, which are depicted in Fig.~\ref{fig:distribution_walks}.

From those figures, we can observe that the distribution of properties in the walks extracted from the enriched graphs is drastically different from those on the original graphs; the Pearson correlation of the distribution in the enriched and original case is 0.44 in the case of walks of depth 4, and only 0.21 in the case of walks of depth 8. The property distributions among the two enrichment strategies, on the other hand, is very similar, with the respective distributions exposing a Pearson correlation of more than 0.99.

Another observation from the graphs is that the distribution is much more uneven for the walks extracted from the enriched graphs, with the most frequent properties being present in the walks at a rate of 14-18\%, whereas the most frequent property has a rate of about 3\% in the original walks. The three most prominent properties in the enriched case -- \emph{location}, \emph{country}, and \emph{locationcountry} -- altogether occur in about 20\% of the walks in the depth 4 setup, and even 30\% of the walks in the depth 8 setup. This means that information related to locations is over-represented in walks extracted from the enriched graphs. As a consequence, the embeddings tend to focus on location-related information much more. This observation might be a possible explanation for the degradation in results on the music and movies datasets being more drastic than, e.g., on the cities dataset.

%For the different datasets, we took a closer look: we computed the property distribution for all the walks both in the materialized and the unmaterialized case, and again calculated the Pearson correlation. We can observe that there is only a low correlation of the property frequencies on the materialized and the unmaterialized graphs. In other words: the property distribution in the walks extracted from the materialized graphs differs significantly from that extracted from the unmaterialized graphs. %The details are depicted in Table~\ref{tab:frequency_correlation}. 

%Contrasting those numbers with the empirical results above, we can observe that, in general, a stronger deviation in the distributions leads to a stronger loss in the resulting models. For example, for the regression datasets, the largest loss was observed for the MetacriticAlbums dataset, where the property distributions in the resulting walks have the lowest correlation with the property distributions in the walks on the unmaterialized graphs.

Finally, we also looked into the correctness of the A-box axioms added. To that end, we sampled 100 axioms added with each of the two enrichment approaches, and had them manually annotated as \emph{true} or \emph{false} by two annotators. For the Wikidata set, the estimated precision is 65.5\% (at a Cohen's Kappa of 0.413), for the DL-Learner dataset, the estimated precision is 61.5\% (at a Cohen's Kappa of 0.73). This shows that the majority of the axioms added to DBpedia are actually correct. Hence, we conclude that a potential addition of erroneous axioms does \emph{not} explain the degradation in the downstream tasks.

%\begin{table*}[t]
%    \caption{Pearson Correlation of the Property Frequencies between Walks Extracted from the Original and the Materialized Graphs for the Different Datasets}
%    \label{tab:frequency_correlation}
%    \bigskip
%    \centering
%    \begin{tabular}{l|l|r|r}
%	&		&	Correlation 4-depth	&	Correlation 8-depth\\
%	\hline
%Wikidata	&	AAUP	&	0.449	&	0.266\\
%	&	CitiesQualityOfLiving	&	0.442	&	0.256\\
%	&	Forbes2013	&	0.341	&	0.261\\
%	&	MetacriticAlbums	&	0.170	&	0.028\\
%	&	MetacriticMovies	&	0.296	&	0.144\\
%	&	KORE	&	0.392	&	0.232\\
%	&	LP50	&	0.405	&	0.220\\
%	\hline
%DL Learner	&	AAUP	&	0.452	&	0.268\\
%	&	CitiesQualityOfLiving	&	0.443	&	0.257\\
%	&	Forbes2013	&	0.348	&	0.267\\
%	&	MetacriticAlbums	&	0.168	&	0.028\\
%	&	MetacriticMovies	&	0.294	&	0.145\\
%	&	KORE	&	0.394	&	0.235\\
%	&	LP50	&	0.406	&	0.222\\
%
%    \end{tabular}
%\end{table*}

\section{Discussion: Missing Information -- Signal or Defect?}
\label{sec:discussion}
Since the results show that adding missing knowledge to the knowledge graph actually results in worse RDF2vec embeddings, we want to investigate the characteristics of missing knowledge in DBpedia in general, as well as its impact on RDF2vec and other algorithms.

\subsection{Nature of Missing Information in Knowledge Graphs}
One first observation is that information in DBpedia and other knowledge graphs is not missing at random. For a curated knowledge graph, a statement is contained in the knowledge graph because some person deemed it relevant.\footnote{For the sake of this argument, we can also consider DBpedia a curated knowledge graph, since the source it is created from, i.e., the infoboxes in Wikipedia, is curated. A statement is contained in DBpedia if and only if somebody considers it relevant enough to be added to an infobox in Wikipedia.}

Consider, e.g., the relation \emph{spouse}. It is unarguably symmetric, nevertheless, in DBpedia, only 9.8k spouse relations are present in both directions, whereas 18.1k only exist in one direction. Hence, the relation is notoriously incomplete, and a knowledge graph completion approach exploiting the symmetry of the \emph{spouse} relation could directly add 18.1k missing axioms.

One example of a \emph{spouse} relation that only exists in one direction is
\begin{verbatim}
Ayda_Field spouse Robbie_Williams .
\end{verbatim}
Ayda Field is mainly known for being the wife of Robbie Williams, while Robbie Williams is mostly known as a musician. This is encoded by having the relation represented in one direction, but not the other. By adding the reverse edge, we cancel out the information that the original statement is more important than its inverse.

Adding inverse relations may have a similar effect. One example in our dataset is the completion of doctoral advisors and students by exploiting the inverse relationship between the two. For example, the fact
\begin{verbatim}
Georg_Joachim_Rheticus doctoralAdvisor 
   Nicolaus_Copernicus .
\end{verbatim}
is contained in DBpedia, while its inverse
\begin{verbatim}
Nicolaus_Copernicus doctoralStudent 
   Georg_Joachim_Rheticus . 
\end{verbatim}
is not (since Nicolaus Copernicus is mainly known for other achievements). Adding the inverse statement makes the random walks equally focus on the more important statements about Nicolaus Copernicus and the ones considered less relevant.

The transitive property adding most axioms to the A-box is the \emph{isPartOf} relation. For example, chains of geographic containment relations are usually materialized, e.g., two cities in a country being part of a region, a state, etc. ultimately also being part of that country. For once, this under-emphasizes differences between those cities by adding a statement making them more equal. Moreover, there usually is a direct relation (e.g., \emph{country}) expressing this in a more concise way, so that the information added is also redundant.

\subsection{Impact on RDF2vec and Other Algorithms}
RDF2vec creates random walks on the graph, and uses those to derive features. Assuming that all statements in the knowledge graph are there because they were considered relevant, each walk encodes a combination of statements which were considered relevant.

If missing information is added to the graph which was \emph{not} considered to be relevant, there is a number of effects. First, the set of random walks encodes a mix of pieces of information which are relevant and pieces of information which are not relevant. Moreover, since the number of walks in RDF2vec is restricted by an upper bound, adding irrelevant information also lowers the likelihood of relevant information being reflected in a random walk. The later representation learning will then focus on representing relevant and irrelevant information alike, and, ultimately, creates an embedding which works worse.

The effects are not limited to RDF2vec. Translational embedding approaches are likely to expose a similar behavior, since they will include both relevant and irrelevant statements in their optimization target, which is likely to cause a worse embedding.

There are also other fields than embeddings where missing information might actually be a valuable signal. Consider, for example, a movie recommender system which recommends movies based on actors that played in the movies. DBpedia and other similar knowledge graphs typically contain the most relevant actors for a movie.\footnote{On average, a movie in DBpedia is connected to 3.7 actors.} If we were able to complete this relation and add all actors even for minor roles, it would be likely that movie recommendations were created on major and minor roles alike -- which are likely to be worse recommendations.

\section{Conclusion and Outlook}
\label{sec:conclusion}
In this paper, we have studied the effect of A-box materialization on knowledge graph embeddings created with RDF2vec. The empirical results show that in many cases, such a materialization has a negative effect on downstream applications.

Following up on those observations, we propose a different view on knowledge graph incompleteness. While mostly seen as a \emph{defect} -- i.e., a knowledge graph is incomplete and hence needs to be fixed -- we suggest that such an incompleteness can also be a \emph{signal}. Although certain axioms could be completed by logical inference, they might have been left out intentionally, since the creators of the knowledge graph considered them less relevant.

A natural future step would be to conduct such experiments on other embedding methods as well. While there is a certain rationale that similar effects can be observed on, e.g., translational embeddings as well, empirical evidence is still outstanding.

Overall, this paper has shown and discussed a somewhat unexpected finding, i.e., that materialization an A-box can actually do harm on downstream tasks, and looked at various possible explanations for that observation.

%\subsubsection*{Acknowledgements}

%The authors would like to thank Frank van Harmelen, who, in a conversation over a beer, first dropped the rumour of a possible degradation of the performance of RDF2vec on materialized graphs.

\bibliographystyle{alpha} 
\bibliography{references}
%inline the .bbl file directly for mailing to authors.

\end{document}

%% file: table_classification_regression.tex
\begin{sidewaystable*}
\caption{Results for Regression (Root Mean Squared Error). \emph{w} stands for number of walks, \emph{d} stands for depth of walks, \emph{v} stands for dimensionality of the RDF2vec embedding space.}
\bigskip
\label{tab:regression}
\scriptsize
\begin{tabular}{l||r|r|r||r|r|r||r|r|r||r|r|r||r|r|r}
	&	\multicolumn{3}{c||}{AAUP}	&		\multicolumn{3}{c||}{CitiesQualityOfLiving} &	\multicolumn{3}{c||}{Forbes2013}	&		\multicolumn{3}{c||}{MetacriticAlbums}	&		\multicolumn{3}{c}{MetacriticMovies}	\\
	Model / Regressor &	LR & KNN & M5 &	LR & KNN & M5 &	LR & KNN & M5 &	LR & KNN & M5 &	LR & KNN & M5 \\
	\hline
500w\_4d\_200v	&	\textbf{67.215}	&	85.662	&	101.163	&	38.364	&	\textbf{14.227}	&	24.271	&	37.509	&	38.846	&	50.411	&	\textbf{11.836}	&	12.110	&	17.414	&	\textbf{20.102}	&	23.888	&	29.901\\
500w\_4d\_200v\_Wikidata  &	70.682	&	82.103	&	105.270	&	47.799	&	15.862	&	24.490	&	\textbf{36.456}	&	37.960	&	51.003	&	13.086	&	13.930	&	18.509	&	21.239	&	23.911	&	30.419\\
500w\_4d\_200v\_dllearner	&	70.340	&	81.991	&	105.403	&	33.326	&	14.931	&	23.629	&	36.602	&	38.504	&	51.298	&	12.997	&	13.973	&	18.573	&	21.402	&	24.102	&	30.506\\
\hline
500w\_4d\_500v	&	\textbf{92.301}	&	95.550	&	103.197	&	15.696	&	15.750	&	26.196	&	43.440	&	39.468	&	51.719	&	13.789	&	\textbf{12.422}	&	17.643	&	\textbf{21.911}	&	26.420	&	30.093\\
500w\_4d\_500v\_Wikidata  &	93.715	&	94.231	&	105.669	&	15.168	&	17.552	&	24.702	&	43.773	&	38.511	&	51.860	&	14.835	&	13.713	&  18.663	&	23.895	&	24.188	&	30.816\\
500w\_4d\_500v\_dllearner	&	92.800	&	97.659	&	106.781	&	\textbf{14.594}	&	16.548	&	25.063	&	43.794	&	\textbf{38.482}	&	52.783	&	14.928	&	13.934	&	18.803	&	23.882	&	24.819	&	30.459\\
\hline
500w\_8d\_200v	&	\textbf{69.066}	&	80.632	&	104.047	&	34.320	&	\textbf{13.409}	&	24.235	&	37.778	&	39.751	&	50.285	&	\textbf{12.237}	&	12.614	&	17.263	&	\textbf{21.353}	&	24.445	&	30.749\\
500w\_8d\_200v\_Wikidata  &	74.184	&	87.009	&	108.335	&	31.482	&	16.124	&	25.706	&	37.588	&	37.985	&	52.294	&	14.028	&	15.340	&	19.415	&	22.456	&	26.002	&	31.597\\
500w\_8d\_200v\_dllearner	&	73.959	&	83.138	&	104.543	&	31.929	&	16.644	&	24.903	&	\textbf{37.212}	&	39.178	&	53.367	&	14.160	&	14.792	&	19.283	&	22.496	&	25.542	&	31.337\\
\hline
500w\_8d\_500v	&	\textbf{92.002}	&	94.696	&	104.326	&	\textbf{11.874}	&	14.647	&	24.076	&	45.568	&	40.827	&	50.976	&	14.013	&	\textbf{12.824}	&	17.579	&	\textbf{23.126}	&	25.146	&	30.457\\
500w\_8d\_500v\_Wikidata  &	97.390	&	104.222	&	108.915	&	15.118	&	17.431	&	26.322	&	44.678	&	\textbf{39.864}	&	50.962	&	16.456	&	15.114	&	19.527	&	25.127	&	26.274	&	31.523\\
500w\_8d\_500v\_dllearner	&	95.408	&	99.934	&	106.267	&	15.055	&	17.695	&	23.680	&	44.516	&	40.647	&	50.060	&	16.260	&	15.131	&	19.458	&	24.396	&	26.127	&	31.397	
\end{tabular}

\caption{Results for Classification (Accuracy). \emph{w} stands for number of walks, \emph{d} stands for depth of walks, \emph{v} stands for dimensionality of the RDF2vec embedding space.}
\bigskip
\label{tab:resultsClassification}
\scriptsize
\begin{tabular}{l||r|r|r|r||r|r|r|r||r|r|r|r||r|r|r|r||r|r|r|r}
	&	\multicolumn{4}{c||}{AAUP}	&		\multicolumn{4}{c||}{CitiesQualityOfLiving} &	\multicolumn{4}{c||}{Forbes2013}	&		\multicolumn{4}{c||}{MetacriticAlbums}	&		\multicolumn{4}{c}{MetacriticMovies}	\\
	Model / Classifier &	NB 	&	KNN	&	SVM	&	C4.5	&	NB 	&	KNN	&	SVM	&	C4.5	&	NB 	&	KNN	&	SVM	&	C4.5	&	NB 	&	KNN	&	SVM	&	C4.5	&	NB 	&	KNN	&	SVM	&	C4.5 \\
\hline
500w\_4d\_200v	&	.564	&	.564	&	\textbf{.659}	&	.526	&	.769	&	.690	&	\textbf{.807}	&	.506	&	.514	&	.519	&   \textbf{.612}	&	.491	&	.723	&	.739	&	\textbf{.764}	&	.612	&	.693	&	.585	&	\textbf{.728}	&	.568\\
500w\_4d\_200v\_Wikidata	&	.607	&	.520	&	.635	&	.490	&	.769	&	.633	&	.798	&	.489	&	.503	&	.508	&	.588	&	.493	&	.662	&	.651	&	.701	&	.558	&	.670	&	.585	&	.681	&	.553\\
500w\_4d\_200v\_dllearner	&	.599	&	.502	&	.626	&	.489	&	.789	&	.715	&	.797	&	.566	&	.518	&	.503	&	.575	&	.490	&	.659	&	.647	&	.688	&	.563	&	.660	&	.582	&	.676	&	.556\\
\hline
500w\_4d\_500v	&	.547	&	.521	&	\textbf{.670}	&	.501	&	.755	&	.596	&	\textbf{.814}	&	.491	&	.496	&	.498	&	\textbf{.606}	&	.497	&	.719	&	.729	&	\textbf{.766}	&	.606	&	.695	&	.531	&	\textbf{.728}	&	.565\\
500w\_4d\_500v\_Wikidata	&	.604	&	.375	&	.641	&	.486	&	.764	&	.555	&	.811	&	.536	&	.507	&	.501	&	.582	&	.485	&	.667	&	.648	&	.705	&	.568	&	.671	&	.527	&	.674	&	.554\\
500w\_4d\_500v\_dllearner	&	.600	&	.298	&	.651	&	.486	&	.722	&	.634	&	.805	&	.512	&	.502	&	.495	&	.567	&	.484	&	.665	&	.635	&	.701	&	.549	&	.672	&	.532	&	.677	&	.558\\
\hline
500w\_8d\_200v	&	.588	&	.589	&	\textbf{.629}	&	.498	&	.791	&	.740	&	.789	&	.530	&	.517	&	.507	&	\textbf{.603}	&	.486	&	.712	&	.726	&	\textbf{.745}	&	.605	&	.676	&	.595	&	\textbf{.692}	&	.556\\
500w\_8d\_200v\_Wikidata	&	.569	&	.485	&	.607	&	.477	&	.736	&	.663	&	\textbf{.808}	&	.522	&	.512	&	.498	&	.576	&	.488	&	.597	&	.546	&	.624	&	.521	&	.630	&	.527	&	.632	&	.528\\
500w\_8d\_200v\_dllearner	&	.588	&	.484	&	.617	&	.481	&	.734	&	.637	&	.800	&	.556	&	.510	&	.494	&	.572	&	.487	&	.616	&	.566	&	.628	&	.530	&	.629	&	.531	&	.634	&	.532\\
\hline
500w\_8d\_500v	&	.599	&	.463	&	\textbf{.658}	&	.510	&	.783	&	.709	&	\textbf{.838}	&	.582	&	.512	&	.490	&	\textbf{.611}	&	.489	&	.699	&	.703	&	\textbf{.739}	&	.605	&	.695	&	.540	&	\textbf{.709}	&	.553\\
500w\_8d\_500v\_Wikidata	&	.566	&	.299	&	.603	&	.470	&	.709	&	.583	&	.815	&	.476	&	.500	&	.493	&	.566	&	.484	&	.574	&	.538	&	.594	&	.520	&	.618	&	.500	&	.631	&	.519\\
500w\_8d\_500v\_dllearner	&	.574	&	.355	&	.598	&	.482	&	.742	&	.589	&	.819	&	.585	&	.493	&	.477	&	.569	&	.489	&	.594	&	.553	&	.611	&	.525	&	.637	&	.507	&	.530	&	.638
\end{tabular}
\end{sidewaystable*}

%% file: table_similarity.tex
\begin{table*}
    \caption{Results for Entity Similarity (Spearman’s Rank)}
    \label{tab:similarity}
    \scriptsize
    \centering
    \begin{tabular}{l||r|r|r|r|r||r}
Model / Dataset	&	IT Companies	&	Celebrities	&	TV Series	&	Video Games	&	Chuck Norris	&	All 21 Entities\\
\hline
500w\_4d\_200v	&	\textbf{.745}	&	\textbf{.702}	&	.586	&	.709	&	\textbf{.540}	&	\textbf{.679}\\
500w\_4d\_200v\_Wikidata	&	.617	&	.503	&	\textbf{.587}	&	.643	&	.448	&	.581\\
500w\_4d\_200v\_dllearner	&	.625	&	.572	&	.574	&	\textbf{.735}	&	.386	&	.615\\
\hline
500w\_4d\_500v	&	\textbf{.720}	&	\textbf{.672}	&	\textbf{.596}	&	\textbf{.753}	&	\textbf{.534}	&	\textbf{.678}\\
500w\_4d\_500v\_Wikidata	&	.603	&	.584	&	.571	&	.668	&	.453	&	.599\\
500w\_4d\_500v\_dllearner	&	.663	&	.581	&	.595	&	.682	&	.469	&	.623\\
\hline
500w\_8d\_200v	&	\textbf{.709}	&	\textbf{.655}	&	\textbf{.539}	&	.681	&	.592	&	\textbf{.643}\\
500w\_8d\_200v\_Wikidata	&	.608	&	.533	&	.448	&	.664	&	\textbf{.603}	&	.565\\
500w\_8d\_200v\_dllearner	&	.632	&	.345	&	.462	&	\textbf{.713}	&	.580	&	.540\\
\hline
500w\_8d\_500v	&	\textbf{.710}	&	\textbf{.693}	&	\textbf{.544}	&	\textbf{.695}	&	\textbf{.710}	&	\textbf{.663}\\
500w\_8d\_500v\_Wikidata	&	.511	&	.509	&	.474	&	.626	&	.513	&	.529\\
500w\_8d\_500v\_dllearner	&	.571	&	.428	&	.517	&	.692	&	.511	&	.550
    \end{tabular}

    \bigskip
    \caption{Results for Entity Relatedness (Spearman’s Rank)}
    \label{tab:relatedness}
    \scriptsize
    \centering
    \begin{tabular}{l||r|r|r|r|r||r}
Model / Dataset	&	IT Companies	&	Celebrities	&	TV Series	&	Video Games	&	Chuck Norris	&	All 21 Entities\\
\hline
500w\_4d\_200v	&	\textbf{.739}	&	\textbf{.651}	&	\textbf{.653}	&	.632	&	.505	&	\textbf{.661}\\
500w\_4d\_200v\_Wikidata	&	.706	&	.508	&	.624	&	.595	&	\textbf{.558}	&	.606\\
500w\_4d\_200v\_dllearner	&	.718	&	.558	&	.582	&	\textbf{.680}	&	.287	&	.618\\
\hline
500w\_4d\_500v	&	\textbf{.749}	&	\textbf{.585}	&	\textbf{.695}	&	.651	&	\textbf{.496}	&	\textbf{.662}\\
500w\_4d\_500v\_Wikidata	&	.696	&	.582	&	.617	&	.590	&	.462	&	.613\\
500w\_4d\_500v\_dllearner	&	.740	&	.578	&	.625	&	\textbf{.695}	&	.386	&	.647\\
\hline
500w\_8d\_200v	&	\textbf{.725}	&	\textbf{.597}	&	\textbf{.629}	&	.593	&	.502	&	\textbf{.630}\\
500w\_8d\_200v\_Wikidata	&	.653	&	.470	&	.514	&	.547	&	\textbf{.711}	&	.554\\
500w\_8d\_200v\_dllearner	&	.690	&	.436	&	.489	&	\textbf{.633}	&	.558	&	.562\\
\hline
500w\_8d\_500v	&	\textbf{.736}	&	\textbf{.634}	&	\textbf{.659}	&	.639	&	.538	&	\textbf{.661}\\
500w\_8d\_500v\_Wikidata	&	.601	&	.406	&	.585	&	.611	&	\textbf{.719}	&	.559\\
500w\_8d\_500v\_dllearner	&	.678	&	.343	&	.509	&	\textbf{.681}	&	.623	&	.556
    \end{tabular}

\end{table*}

%% file: table_document.tex
\begin{table*}[t]
\caption{Results for the Document Similarity Task. \emph{w} stands for number of walks, \emph{d} stands for depth of walks, \emph{v} stands for dimensionality of the RDF2vec embedding space.}
\scriptsize
\label{tab:document_similarity}
\centering
\begin{tabular}{l|r|r|r}
Model / Metric	&	Pearson Score	&	Spearman Score	&	Harmonic Mean\\
\hline
500w\_4d\_200v	&	.241	&	.144	&	.180\\
500w\_4d\_200v\_Wikidata	&	.146	&	.161	&	.154\\
500w\_4d\_200v\_dllearner	&	\textbf{.252}	&	\textbf{.190}	&	\textbf{.217}\\
\hline
500w\_4d\_500v	&	.105	&	.015	&	.027\\
500w\_4d\_500v\_Wikidata	&	.073	&	\textbf{.086}	&	.079\\
500w\_4d\_500v\_dllearner	&	\textbf{.116}	&	\textbf{.086}	&	\textbf{.099}\\
\hline
500w\_8d\_200v	&	.231	&	.192	&	.210\\
500w\_8d\_200v\_Wikidata	&	.242	&	\textbf{.227}	&	.234\\
500w\_8d\_200v\_dllearner	&	\textbf{.315}	&	\textbf{.227}	&	\textbf{.264}\\
\hline
500w\_8d\_500v	&	.196	&	.174	&	.185\\
500w\_8d\_500v\_Wikidata	&	.193	&	.175	&	.184\\
500w\_8d\_500v\_dllearner	&	\textbf{.238}	&	\textbf{.192}	&	\textbf{.213}
\end{tabular}
\end{table*}